# Data Issues in Industrial AI System: A Meta-Review and Research Strategy

Xuejiao Li[1,3,*], Yang Cheng[1], Charles Møller[2], Jay Lee[3]

1: Center for Industrial Production, Department of Materials and Production, Aalborg University, Aalborg, Denmark

2: Department of Mechanical and Production Engineering, Aarhus University, Denmark

3: Industrial Artificial Intelligence Center, Department of Mechanical Engineering, University of Maryland, MD, USA

*: Corresponding author, xueli@mp.aau.dk

**Abstract:**

In the era of Industry 4.0, artificial intelligence (AI) is assuming an increasingly pivotal role within industrial systems. Despite the recent trend within various industries to adopt AI, the actual adoption of AI is not as developed as perceived. A significant factor contributing to this lag is the data issues in AI implementation. How to address these data issues stands as a significant concern confronting both industry and academia. To address data issues, the first step involves mapping out these issues. Therefore, this study conducts a meta-review to explore data issues and methods within the implementation of industrial AI. Seventy-two data issues are identified and categorized into various stages of the data lifecycle, including data source and collection, data access and storage, data integration and interoperation, data pre-processing, data processing, data security and privacy, and AI technology adoption. Subsequently, the study analyzes the data requirements of various AI algorithms. Building on the aforementioned analyses, it proposes a data management framework, addressing how data issues can be systematically resolved at every stage of the data lifecycle. Finally, the study highlights future research directions. In doing so, this study enriches the existing body of knowledge and provides guidelines for professionals navigating the complex landscape of achieving data usability and usefulness in industrial AI.

**Keywords:** Data issue; Data lifecycle; AI; Industrial AI; Machine learning

# 1. Introduction:

In the era of Industry 4.0, artificial intelligence (AI) is assuming an increasingly pivotal role within industrial systems. Consequently, the discussion on industrial AI draws increasing attention. Industrial AI can be defined as a systematic discipline to enable engineers to systematically develop and deploy AI algorithms with repeating and consistent successes (Lee et al., 2019). It focuses on the development, validation, deployment, and maintenance of AI solutions for industrial applications with sustainable performance (Lee, 2020). However, AI algorithms cannot stand alone. When discussing AI, the specific requirements of the industrial application and the available data is inevitable. The relationship between data, AI, and industrial applications is symbiotic. Data serves as the foundation upon which AI algorithms operate within industrial systems. Through AI, industrial applications can harness the power of data to make data-driven decisions. In turn, industrial applications generate vast amounts of data that fuel further AI development and refinement. This cyclical relationship enables continuous improvement and innovation in industrial processes, leading to increased efficiency, productivity, and competitiveness.

The flow from data to AI to industry is depicted in Figure 1. Once the decision to implement AI is made, the flow starts with data. This involves collecting relevant data, storing it appropriately, ensuring data quality through cleaning, and analyzing and visualizing data for the intended purposes. Following this, AI models are brought into consideration. The process begins with extracting features from the data by feature engineering, then choose a suitable AI model based on the nature of the problem and data. Afterwards, the model needs to be trained and validated to ensure effective generalization, and the performance of the model needs to be evaluated using appropriate metrics. Finally, the AI model can be implemented in the industrial setting. The application areas in the industries encompass process optimization, predictive maintenance, quality control, collaborative robotics, ergonomics, and workforce training (Peres et al., 2020). Industrial sectors that benefit from the adoption of AI include embedded AI devices, resilient factories, smart human and health performance, predictive energy systems, worry-free transportation, and industrial AI-based education systems (Lee et al., 2018).

Along with the flow, knowledge and experience, such as data sources selection, data preprocessing steps, and model deployment details, can be documented and accumulated. The knowledge and experience drive continuous improvement by refining data processes, optimizing AI algorithms, enhancing industrial processes, integrating emerging technologies, fostering organizational learning, and enabling adaptation to changing conditions, etc. In the end, the interconnection between data, AI, and industrial systems forms a closed loop by continuous improvement.

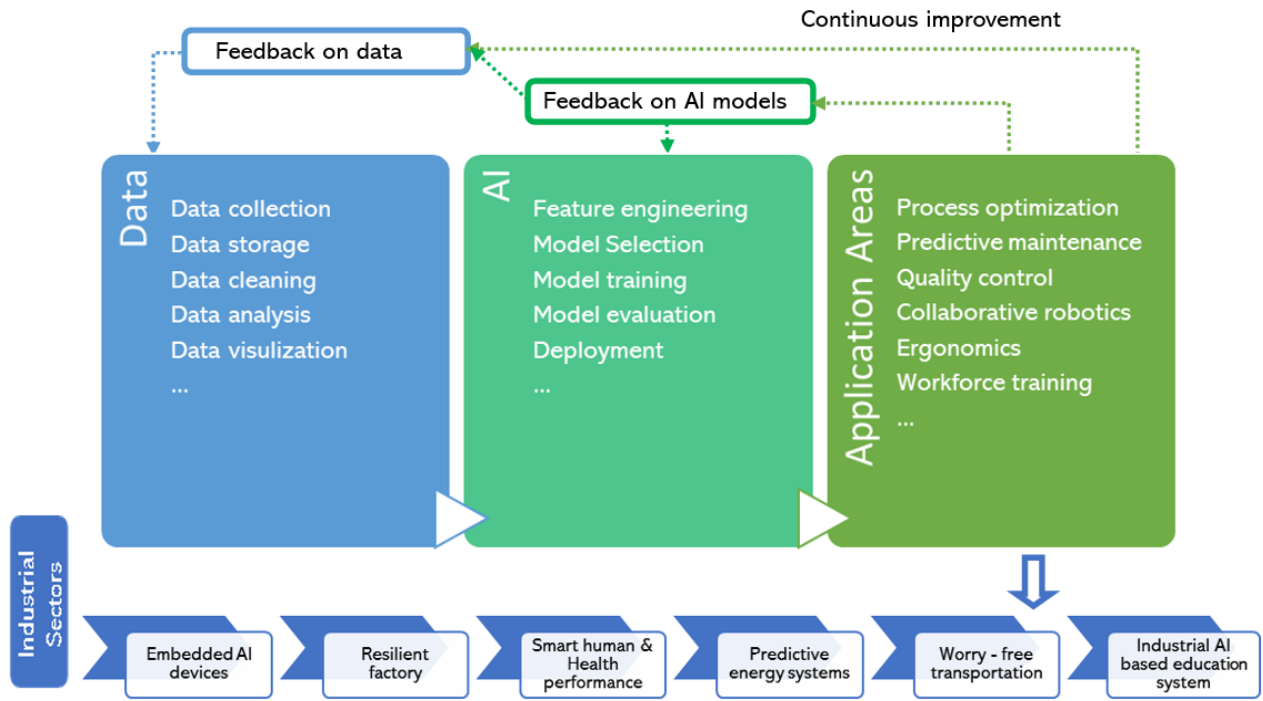

Figure 1. Closed loop of data, AI, and industrial applications

In the closed loop, data serves as the lifeblood. The collection, preparation and analysis of data plays a crucial role in determining the success of AI initiatives. As industries become more data-driven, the critical importance of ensuring the usability and usefulness of data cannot be overstated. However, in practice, numerous issues are linked to data. These include, but are not limited to, insufficient data, imbalanced data, data fusion issues, and data privacy and security concerns. How to address these data issues stands as a significant concern confronting both industry and academia. Hence, this study posits that the primary step in progressing the implementation of industrial AI is to systematically tackle the data issues along with their corresponding methods. To do so, this study aims to explore the following research questions:

**RQ1. What specific data issues arise during the implementation of AI in industrial systems?**

**RQ2. What methods are associated with addressing these issues?**

**RQ3. How can these data issues be systematically resolved?**

Aligned with this rationale, this study undertakes a meta-review of existing reviews on data issues and methods in the context of AI implementation in industrial systems. Subsequently, it categorizes these issues into data lifecycle stages to systematically resolve them. Building upon the results, it then presents a framework for comprehensively handling data issues in industrial AI.

This study contributes to the existing body of knowledge and holds significance in three dimensions. First, it identifies gaps within the current research landscape of industrial AI and offers suggestions for future research. Second, it outlines data issues, methods, and best practices to offer practical guidelines. Third, it provides a comprehensive data management framework to systematically handle data issues at every stage of the data lifecycle. These guidelines and the framework can be valuable resources for professionals navigating the complex terrain of achieving data usability and usefulness in industrial AI implementation.

The rest of this study is organized as follows: Section 2 compares the research that aligns with similar interests with our paper and provides background on the data lifecycle theory. Section 3 introduces the methodology of how we collect, extract, and analyze data. Section 4 presents the results of the literature analysis. Section 5 proposes a data management framework for systematically handling data issues in industrial AI. Section 6 identifies the future research directions on data forum in industrial AI. Finally, conclusions and limitations are described in Section 7.

**2. Literature review**

Systematic literature reviews are regarded as one of the most rigorous forms of scientific evidence, providing a robust method for identifying and synthesizing existing literature (Mallett et al., 2012; Matricciani et al., 2019). Numerous articles have mentioned data in AI within the industrial systems, with many of them being review articles. Among these reviews, some share our interest in taking data as the main research object. While these reviews offer valuable insights, they differ in the research focus and scope compared to our study.

A literature review paper (Zhang et al., 2023b) examines publications on the topic of data quality in additive manufacturing (AM). It concentrates on various data types, including tabular, graphics, 3D, and spectrum data, as well as the corresponding data handling methods such as feature extraction, feature selection and learning, discretization, and data preprocessing. This paper further discusses the implementation of machine learning (ML) algorithms in AM. It also introduces existing public AM databases and data management methods. However, it does not focus on specific data issues and the research scope is restricted solely on the domain of AM.

Another review (Zhang & Gao, 2021) focuses on two primary aspects: data curation, aimed at providing high-quality data for meaningful deep learning (DL)-based analysis, and model interpretation. This review summarizes key techniques in data curation, including data denoising, outlier detection, imputation, balancing, and semantic annotation, and also showcases their effectiveness in extracting information from noisy, incomplete, insufficient, and/or unannotated data. While the discussed topics of data issues, data curation techniques, and model interpretation methods are highly relevant to our study, it exclusively focuses on DL, which represents only a subfield within the broader scope of AI technology.

Baloch et al. (2023) explored the challenges and opportunities associated with employing big data in healthcare, including data security, privacy, data quality, interoperability, and ethical considerations. They also examined the potential applications of big data, encompassing personalized treatment, disease prediction and prevention, and population health management. The study offers valuable insights for healthcare providers, researchers, and practitioners. However, this review is limited to the healthcare domain.

The work of Madhikermi et al. (2016) is a thorough literature review of both existing expert maintenance systems making use of Multi-Criteria Decision Making (MCDM) techniques, and existing data quality frameworks. The criteria related to the data quality dimensions, as discussed in this study, such as believability, completeness, and timeliness, is helpful for us to better understand data quality. The Krogstie's data quality framework mentioned in this study including physical quality, syntactic quality, semantic quality, pragmatic quality, social quality, knowledge quality, and language quality is also inspiring. However, this study does not delve into specific data issues. In addition, it only focuses on the data quality assessment for maintenance reporting procedures.

Arruda et al. (2023) conducted a systematic literature review to explore the application of data science methods and tools across various industrial segments. It considers different time series levels and data quality. This study identifies 16 industrial segments, 168 data science methods, and 95 software tools, presenting the findings in a taxonomic approach for a comprehensive state-of-the-art representation and visualization. While this study is valuable for summarizing industrial segments and cataloging data science methods and tools used in industrial systems, it is important to note that these methods and tools serve broader applications beyond addressing specific data issues as we aim to do. Additionally, the consideration of data quality in this study is limited to the quantity and origins of datasets, giving only a partial insight into data quality.

In addition, there are two reviews concentrating on techniques which can be used to address data issues. One of them delves into outlier detection in Wireless Sensor Networks (WSNs) (Lee & Su, 2023). It argues that in WSNs, sensor nodes distributed autonomously in harsh environments that cause sensor readings to be unreliable and inaccurate. Sensor readings that have differed considerably from healthy behaviors are considered outliers. Such outliers in data analytics will affect the outcome of the decision-making process. Thus, detecting outliers in WSNs using data-driven approaches becomes a novel technique. This review then presents a comprehensive overview of the state-of-the-art on Statistical and AI based techniques used in WSNs to detect outliers. Another review focuses on feature engineering for data-driven building energy prediction (Wang et al., 2022). This study discusses the concept of feature engineering and its main methods, including the construction, selection, and extraction of features, then summarizes the status and characteristics of feature engineering research in the building energy domain. The study also discusses critical issues in feature engineering and identifies future research directions. Both reviews have a specific focus, offering in-depth insights into outlier detection and feature engineering.

As highlighted previously, while various reviews touch upon data quality and data issues topics in AI applications, none comprehensively address specific data issues and corresponding methods in the industrial systems scope. This study aims to address this gap by conducting a meta-review, a rigorous methodological approach known as "reviews of reviews." Meta-reviews systematically evaluate existing reviews, offering a comprehensive assessment of academic work within a specific field (Ryan et al., 2009). This choice was prompted by the existence of numerous literature reviews that touch on data issues but lack a holistic examination.

Meta-reviews are increasingly favored in situations where conducting systematic reviews becomes challenging due to the extensive volume of literature or when multiple reviews with diverse findings exist. Through this approach, our study serves as a rapid means of mapping past efforts, identifying research gaps, and highlighting areas for improvement. There has been no meta-review addressing data issues in industrial AI systems to date. This study aims to fill this void by synthesizing findings from literature reviews related to data research in the implementation of AI within industrial systems. The scope of our meta-review encompasses diverse areas within industrial systems and various AI technologies.

## 3. Methodology

### 3.1 Study identification, screening, and eligibility

The meta review follows the guidelines specified in the PRISMA statement (Mallett et al., 2012). Figure 1 displays the PRISMA flow chart, illustrating the various phases of the study.

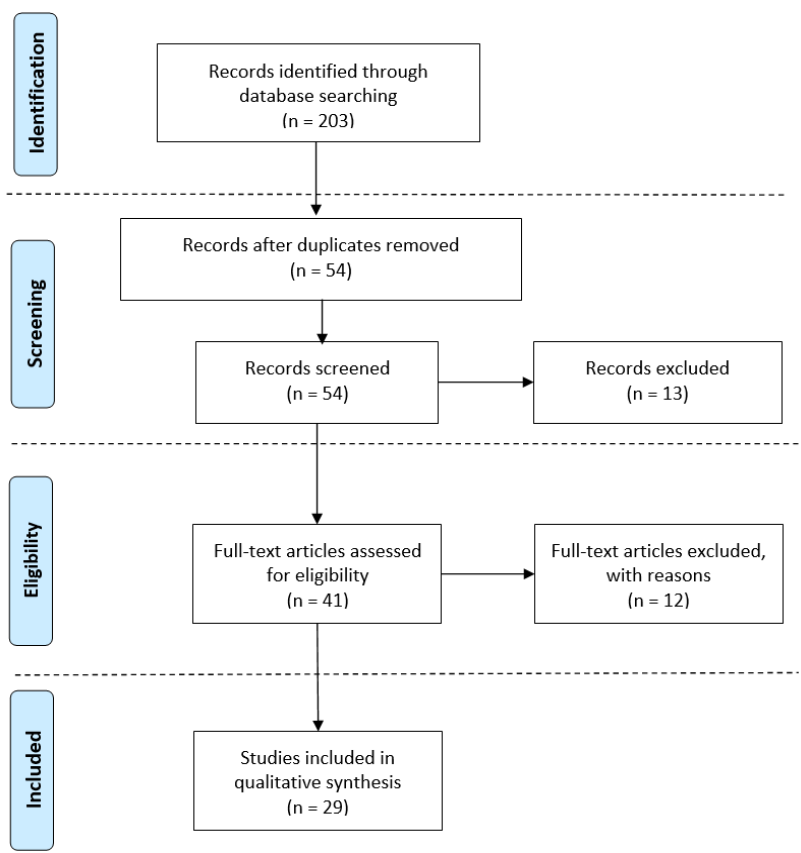

Figure 1. PRISMA flowchart of study inclusions and exclusions for the systematic literature review. Adopted from Moher et al., 2009

Initially, a keyword string was formulated with a focus on data research related to the implementation of AI within industrial systems. The construction involved ensuring the inclusion of at least one element from each aspect listed in Table 1, employing a combination of OR and AND operators. The three aspects encompass data issue, industrial system, and AI. Regarding the first two aspects, the elements in Table 1 represent synonyms or antonyms associated with these aspects. As for AI, the elements in Table 1 constitute the core concepts associated with AI, sourced from reference (Peres et al., 2020).

Table 1. The aspects that the subject contains and the search terms

| Aspect 1 (Data issue) | Aspect 2 (Industrial System) | Aspect 3 (AI) |
|---|---|---|
| Data issue | Industr* | AI |
| Data error | Manufactur* | Artificial intelligence |
| Data problem | Craft | Machine Learning |
| Data quality | Sector | Deep Learning |
|  | Commerce | Data science |
|  |  | Predictive analytics |

Subsequently, this search string was tailored for application across the three electronic databases central to this study: Web of Science, Scopus, and ScienceDirect. The search was executed on December 6, 2023, encompassing academic research that fulfilled the following criteria: (1) Only Peer Reviewed publications; (2) published between 2014 and December 2023 (recent 10 years); (3) Document type: Review; (4) containing at least one term from each group in the abstract, title, or keywords; and (5) composed in the English language. Following this, the obtained records were consolidated, and duplicates were removed.

After the identification phase, two independent reviewers conducted the first screening process based on the exclusion criteria for the screening phase listed in Table 2. Any discrepancies were resolved through discussion.

Table 2. Exclusion criteria for the screening phase

| No. | Screening criteria |
|---|---|
| 1 | The record must be in English. |
| 2 | The record must include at least the Title, Year, Source, Abstract, and DOI. |
| 3 | The record must be a systematic review. |
| 4 | The abstract must discuss data in the context of AI implementation. Any discrepancies were resolved through discussion. |

Finally, the remaining articles underwent a detailed analysis of their full text, based on the exclusion criteria for the eligibility phase listed in Table 3. No further restrictions were applied.

Table 3. Exclusion criteria for the eligibility phase

| No. | Screening criteria |
|---|---|
| 1 | The record is not relevant to data quality or data issues. |
| 2 | The record is not relevant to AI. |
| 3 | The domain of the record is beyond the industrial system. |
| 4 | Full text is not available. |

**3.2 Data extraction**

For each eligible article included in the study, two types of data were extracted. Initially, basic information about the publication was gathered, encompassing (1) source title, (2) publication title, (3) authors, (4) keywords, (5) abstract, (6) year, (7) subject area, and (8) country/territory.

The second part is dedicated to addressing the research questions, for RQ1 "What are the specific data issues encountered during the implementation of AI in industrial systems", the data extracted from the eligible publications are: 1) specific AI technologies; 2) domain with the industrial systems; 3) data issues. For RQ2 "What are the associated methods to these issues", the data extracted from the eligible publications is: 4) methods for addressing data issues. For RQ3 "How can data issues be systematically resolved", the data extracted from the eligible publications is: 5) data lifecycle stage of the data issues. Nevertheless, when it comes to point 5), extracting data from the text in the publications may not always be straightforward. Hence, distinguishing and interpretation during the data analysis phase is essential to discern the corresponding data lifecycle stage.

**3.3 Data analysis**

We use data lifecycle theory for data analysis in this study. The data lifecycle theory provides a systematic approach to understanding, managing, and improving the progression of data throughout its lifecycle. When applied to analyzing data issues, this theory offers a structured framework for comprehensively examining data from inception to disposal. By systematically addressing data issues within this framework, organizations can establish clear guidelines for data management and governance.

The data lifecycle encompasses a sequence of phases covering its entire useful lifespan, typically including various stages. More and more studies explore the phases of the data lifecycle. For example, an influential article by Tao et al. (2018) delineates seven distinct phases within the manufacturing data lifecycle, comprising data sources, data collection, data storage, data processing, data visualization, data transmission, and data application. Another study (Wing, 2019) defines the data lifecycle as involving generation, collection, processing, storage, management, analysis, visualization, and interpretation of data. Ashmore et al. (2021) and Paleyes et al. (2022) specifically emphasize the significance of data management in the implementation of ML. They highlight that data management plays a crucial role in acquiring the data essential for synthesizing ML models. This process encompasses data collection, data preprocessing, data augmentation, and data analysis. However, scholars have not yet reached a consistent consensus regarding the precise phases that constitute the data lifecycle. In this study, with a specific emphasis on AI-related data, we synthesize existing research while specifically addressing the data pipeline in methodologies such as ML. Therefore, we emphasize phases including *data source and collection, data access and storage, data integration and interoperation, data pre-processing, data processing,* and *data security and privacy* in the data lifecycle. Additionally, we observed that apart from data itself, there are also data issues associated with the technologies used in implementing AI. Hence, we have incorporated *AI Technology adoption* as an additional category within our analytical framework.

## 4. Results

This section provides the results of data analysis, which include the statistical description of the selected articles, distribution of AI technologies and industrial domains, and data issues and methods.

### 4.1 Statistical description of selected articles

Understanding the landscape of selected articles is paramount to gaining insights into the prevailing trends, thematic concentrations, and scholarly contributions. This section provides a statistical overview, employing Figures 2, 3, and 4 to offer an examination of the selected literature.

Figure 2 traces the distribution of articles across different publication years, shedding light on the evolving patterns of research activity. Based on the articles we collected, the peak of interest in the domain appears to be in 2023. However, we can only gain a full picture of the number of publications for 2024 in early 2025. Therefore, there is a possibility that the number of publications in 2024 exceeds that of 2023, indicating an increasing trend.

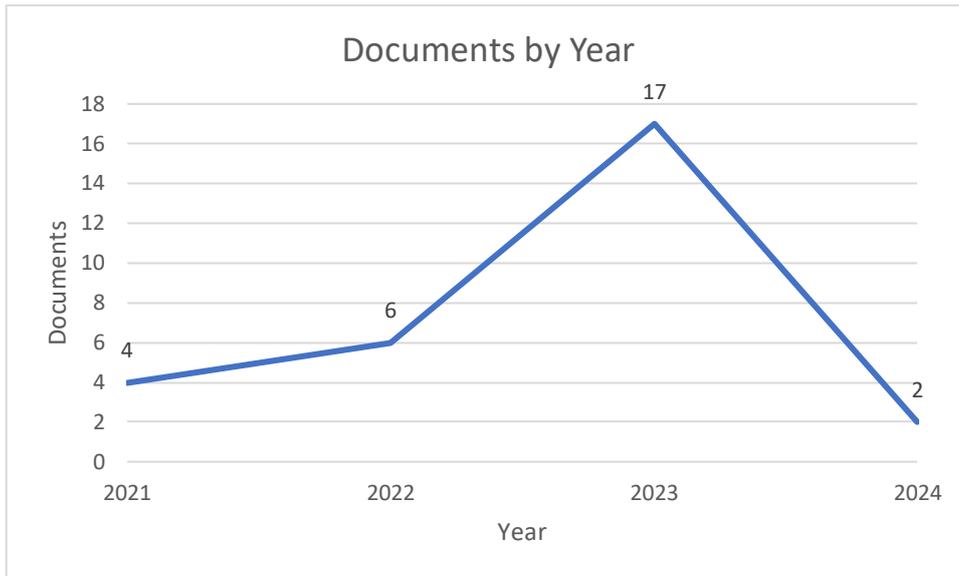

Figure 2. Documents by year

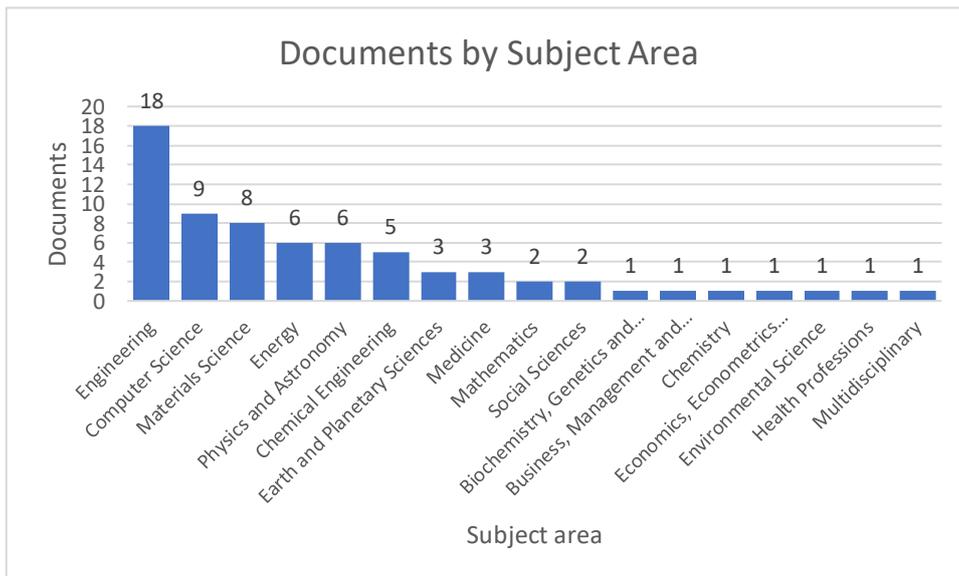

Figure 3. Documents by Subject Area

Figure 3 categorizes articles by subject area, revealing the thematic diversity and concentration. The primary emphasis of most articles lies within the areas of Engineering, Computer Science, and Material Science.

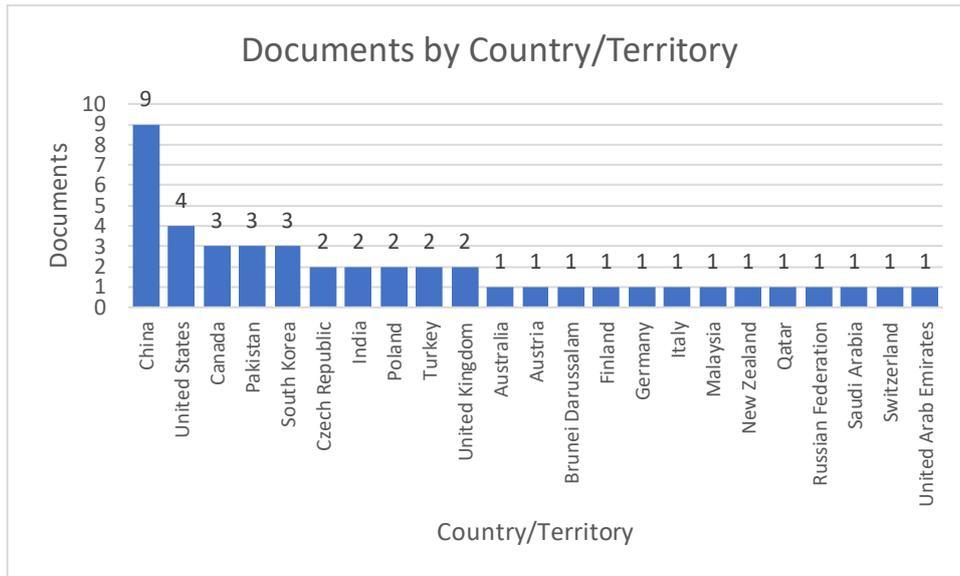

Figure 4. Documents by Country/Territory

Figure 4 categorizes articles by country/territory, showcasing the diversity and concentration in geographical distribution. The primary contributors to selected data issue reviews are observed to be China, the United States, and Canada.

**4.2 Distribution of AI technologies and industrial domains**

Regarding AI technologies and techniques, observing the selected studies, it becomes evident that the majority of articles center around ML, with approximately one-third of them (9 articles) specifically referencing DL, a subfield of ML. Two articles are made to Artificial Neural Networks (ANNs) (Akhtar et al., 2023; Strielkowski et al., 2023a), which are the computational models inspired by the structure and functioning of biological neural networks in the human brain and can be employed in various ML tasks. One reference mentions Natural Language Processing (NLP) (Baloch et al., 2023).

Moreover, certain articles delve into more specific ML and DL techniques. Notably, transformer architectures (Sengupta, 2023), a category of DL model architecture, have showcased superior performance across a broad spectrum of NLP tasks (Ashish, 2017). Other mentioned techniques include ML algorithms such as support vector machines (SVMs), decision trees, and random forests (Strielkowski et al., 2023); Automated Machine Learning (AutoML) (Wen & Li, 2022); Physics-Based Machine Learning (PBML) (Yüce et al., 2022); Supervised Machine Learning (SML) (Roy et al., 2022); Feature engineering (Wang et al., 2022) for enhancing ML model performance; Transfer Learning (Zhu et al., 2022), a technique where a model developed for a specific task serves as the starting point for a model on a second task; Contrastive Learning (Zhu et al., 2022), a self-supervised learning technique in ML and computer vision; and Incremental Learning (Zhu et al., 2022), a ML paradigm where a model undergoes continuous updates and improvements as new data becomes available over time.

Regarding domains in the industrial systems, the selected articles, as depicted in Table 5, span a diverse array of areas. The majority of coverage includes energy and power systems (Liu et al., 2023; Akhtar et al., 2023; Strielkowski et al., 2023a; Strielkowski et al., 2023b; Yüce et al., 2022; Wang et al., 2022; Zhu et al., 2022), healthcare (Cellina et al., 2023; Sengupta, 2023; Baloch et al., 2023; Wong et al., 2023; Roy et al., 2022; Wilson et al., 2021), manufacturing (Zhang et al., 2023b; Xu et al., 2022; Tripathi et al., 2021; Zhang & Gao, 2021), and the material industry (Ma et al., 2023; Guo et al., 2023).

Table 5. AI technologies and industrial domains

| Reference | AI techonology | Domain |
| --- | --- | --- |
| KHALAF et al., 2023 | ML | Oil and gas industry - maintenance - corrosion monitoring |
| Abrasaldo et al., 2024 | ML | Geothermal energy industry - above-ground geothermal operations |
| Durlik et al., 2023 | ML | Maritime industry, maritime operations and maintenance |
| Cellina et al., 2023 | DL | Healthcare industry - application of digital twins, digital human twins |
| Aldoseri et al., 2023 | / | / |
| Sengupta, 2023 | DL, Transformer architectures | Healthcare - dermatology, the diagnostic aspect of dermatological conditions |
| Liu et al., 2023a | ML, DL | Wind power - prediction, forecasting |
| Zhang et al., 2023a | ML | Metrology or precision measurement |
| Liu et al., 2023b | DL | Mining industry, sorting and optimal utilization of minerals, digital mineral image classification |
| Zhang et al., 2023b | ML | AM |
| Ma et al., 2023 | DL | Material industry, material microscopic image analysis |
| Baloch et al., 2023 | NLP, ML, DL | Healthcare sector |
| Wong et al., 2023 | ML | Biomedical research and healthcare |
| Akhtar et al., 2023 | ANNs, ML | Energy industry |
| Strielkowski et al., 2023a | ML, SVMs, decision trees, random forests; ANNs | Power systems sector |
| Wen & Li, 2022 | AutoML | Spatial decision support systems |
| Guo et al., 2023 | ML | Materials science and engineering |

| Shi et al., 2023 | ML | Blast furnace ironmaking |
| Strielkowski et al., 2023b | DL | Electric power systems |
| Yüce et al., 2022 | ML, PBML | Wind energy infrastructure |
| Roy et al., 2022 | SML | Healthcare and biomedical research |
| Xu et al., 2022 | DL | Smart manufacturing |
| Chuo et al., 2022 | Various | Machining operations |
| Wang et al., 2022 | ML, feature engineering | Building energy prediction |
| Zhu et al., 2022 | Transfer Learning, Contrastive Learning, Incremental Learning | Integrated Energy Systems |
| Field et al., 2021 | ML | Radiation oncology |
| Wilson et al., 2021 | / | Healthcare |
| Tripathi et al., 2021 | ML | Production and Manufacturing |
| Zhang & Gao, 2021 | DL | Smart manufacturing |

**4.3 Data issues and methods**

As mentioned before, this study aims to provide a comprehensive overview of existing reviews on data issues and methods in industrial AI implementation. 72 data issues are identified and categorized into various stages of the data lifecycle, along with corresponding methods for addressing these issues. The data lifecycle stages include data source and collection, data access and storage, data integration and interoperation, data pre-processing, data processing, data security and privacy, and AI Technologies adoption. The findings are presented in Tables 6 to 12 as outlined below.

**1) Data Source and Collection**

In the Data Source and Collection stage, a total of 12 data issues have been identified, which can be classified into three types: insufficient data, excessive data, and overly diverse data. Insufficient data implies a lack of necessary information. For instance, challenges in collecting sufficient, accurate, and high-quality data are evident (KHALAF et al., 2023; Sengupta, 2023), along with limitations in data availability (KHALAF et al., 2023; Ma et al., 2023; Wong et al., 2023; Akhtar et al., 2023; Zhu et al., 2022; Wilson et al., 2021). The corresponding methods vary and can be categorized into both management approaches, such as collaborative efforts among developers, startups, and academia for data collection, and technical methods. Technical methods include constructing large open-source datasets that are anonymized, representative, and regularly updated; creating comprehensive benchmarking datasets; employing data augmentation techniques like generative algorithms to artificially generate representative data; transfer learning, and domain adaptation techniques.

Excessive data refers to a situation where data is immense in volume, high in velocity, and diverse in variety (Durlik et al., 2023; Aldoseri et al., 2023). To address this, industry-wide initiatives promoting data standardization can be advocated (Durlik et al., 2023). Additionally, establishing clear data requirements, including the identification of data types, sources, and quantities, is recommended (Aldoseri et al., 2023).

The challenge of overly diverse data sources arises when dealing with varying data formats from different sources (KHALAF et al., 2023), coupled with a lack of standardization (Strielkowski et al., 2023a). While specific methods for addressing overly diverse data sources are not explicitly outlined in the references, one potential approach could involve establishing standards for data sources and collection processes.

Table 6. Data issues and methods in the Data Source and Collection stage

| Data issue | Specific issue | Method | AI technology | Domain |
|---|---|---|---|---|
| **Insufficient data** | 1. Difficult to collect sufficient and accurate data (KHALAF et al., 2023) | / | ML | Oil and gas industry – maintenance |
| | 2. Lack of historical data (KHALAF et al., 2023) | / | ML | Oil and gas industry – maintenance |
| | 3. Difficult to find high-quality, diverse, and representative data (Sengupta, 2023) | Developers, startups, and academia collaborate on data collection; Build large open-source, anonymized, representative, and regularly updated datasets; Develop generative algorithms to artificially create representative images. | Transformer architectures, DL | Healthcare |
| | 4. Limited availability of annotated data for training deep learning models (Ma et al., 2023) | Image Augmentation | DL | Material industry, material microscopic image analysis |
| | 5. Limited data availability and generalizability (Wong et al., 2023) | Comprehensive benchmarking datasets for antigen selection and vaccine efficacy | ML | Biomedical research and healthcare |
| | 6. Lack of data (Akhtar et al., 2023) | Data augmentation techniques, transfer learning and domain adaptation techniques | ML, ANNs | Energy industry |
| | 7. Limited data (Zhu et al., 2022) | Transfer Learning, involves leveraging knowledge from a correlated domain to improve outcomes in the target domain; Generative Models: such as GANs to | Transfer Learning, Contrastive Learning, GANs, | Integrated Energy Systems |

| | | generate synthetic but realistic time-series data, addressing the challenge of limited data on hand | Incremental Learning | |
|---|---|---|---|---|
| | 8. Not all events are captured in the medical record (Wilson et al., 2021) | Data cleaning, artefact removal | / | Healthcare |
| **Excessive data** | 1. Sheer volume and variety of data, immense in volume, high in velocity, and diverse in variety (Durlik et al., 2023) | Champion industry-wide data standardization initiatives | ML | Maritime industry, maritime operations and maintenance |
| | 2. Data deluge, the diversity of data sources, and the need for representative samples (Aldoseri et al., 2023) | Data requirements for identifying data types, sources, and quantities | ML | / |
| **Overly diverse data sources** | 1. Varying data formats from different sources (KHALAF et al., 2023) | / | ML | Oil and gas industry – maintenance |
| | 2. Lack of standardization across different data sources (Strielkowski et al., 2023a) | / | ML | Power systems sector |

## 2) Data Access and Storage

In the Data Access and Storage stage, 2 data issues have been identified, labeled as "no access" and "cost issue." "No access" refers to the inability to utilize large datasets due to legal or regulatory constraints (Liu et al., 2023). The recommended method for addressing this issue involves applying for the necessary permissions and licenses to access and use the data. On the other hand,

Cost issues indicate challenges in storing data while meeting scalability and performance requirements within budget constraints (Aldoseri et al., 2023). Proposed methods to mitigate this challenge include maintaining existing data storage and retrieval systems and employing technologies such as nonvolatile memory or distributed storage systems for storing data.

Table 7. Data issues and methods in the Data Access and Storage stage

| Data issue | Specific issue | Method | AI technology | Domain |
|---|---|---|---|---|
| **No access** | 1. Not able to use large datasets due to legal or regulatory constraints (Liu et al., 2023) | Necessary permissions and licenses to access and use data | ML, DL | Wind power - prediction, forecasting |
| **Cost issue** | 1. Difficult to meet scalability, performance requirement within budget on data storage (Aldoseri et al., 2023) | Maintaining data storage and retrieval systems; Nonvolatile memory; Distributed storage systems | ML | / |

## 3) Data Integration and Interoperation

Within the Data Integration and Interoperation stage, 4 data issues have been identified, characterized as "system issue" and "data fusion issue". System issues entail the lack of integration in existing systems (KHALAF et al., 2023; Wilson et al., 2021), coupled with a deficiency in standardization and interoperation between systems (Durlik et al., 2023). When efforts are made to integrate systems, challenges such as compatibility issues and resistance to change from traditional practices may arise (KHALAF et al., 2023). In some instances, unintegrated systems are still reliant on paper records (Wilson et al., 2021). The methods for addressing these issues can be broadly categorized into management and technical approaches. For management methods, initiatives such as catalyzing joint development and uniting stakeholders, advocating for collaborative data sharing platforms for cross-verification, anticipating challenges in data encoding and extraction, and fostering collaboration with IT teams are recommended. On the technical front, employing advanced data integration techniques, integrating versatile middleware solutions for system harmonization, preparing for extensive data wrangling to align datasets from various sources, and accurately defining project data are suggested.

Regarding the data fusion issues, Zhang et al. (2023a) suggest the presence of potential data quality issues without providing specific details. Gaussian Processes (GPs) algorithms are proposed as a viable solution. GPs, a type of non-parametric model widely used in machine learning for regression and probabilistic classification, create a distribution over functions. This unique characteristic allows them to adeptly capture complex relationships between input and output variables. Yüce et al. (2022) discuss the difficulties in quantifying confidence in fused data and suggests the adoption of a Bayesian framework to construct more precise surrogate models.

Table 8. Data issues and methods in the Data Integration and Interoperation stage

| Data issue | Specific issue | Method | AI technology | Domain |
|---|---|---|---|---|
| **System issue** | 1. Integration with Existing Systems: Compatibility issues, resistance to change from traditional practices (KHALAF et al., 2023) | Develop advanced data integration techniques | ML | Oil and gas industry – maintenance |
| | 2. Lack of standardization and interoperation between systems (Durlik et al., 2023) | Integrate versatile middleware solutions for system harmonization; Catalyze joint development initiatives, uniting stakeholders; Advocate for collaborative data sharing platforms for cross-verification | ML | Maritime industry, maritime operations and maintenance |
| | 3. IT systems are not integrated and are often supported by paper records | Be prepared for extensive data wrangling to align datasets from various sources; Anticipating data | / | Healthcare |

| | (Wilson et al., 2021) | encoding and extraction challenges; Accurately defining project data; Collaboration with IT teams | | |
|---|---|---|---|---|
| **Data fusion issue** | 1. Data quality issues in data fusion (Zhang et al., 2023a) | GPs algorithms | ML | Metrology, precision measure |
| | 2. Quantify the confidence in the fused data (Yüce et al., 2022) | Using a Bayesian framework to construct more accurate surrogate models | ML | Wind energy infrastructure |

**4) Data Pre-processing**

The majority of data issues are concentrated in the data pre-processing stage. A total of 37 issues have been identified, grouped into 13 types. These 13 types can be further divided into two groups. The first group pertains to data quality issues, where data quality is assessed based on criteria such as accuracy, reliability, consistency, completeness, and relevance for a specific purpose. Within this group, the study addresses issues like *incompleteness, imbalance, inaccuracy, outliers, and mislabeling*. The second group focuses on data nature issues, referring to the inherent characteristics or properties of data. Data nature is influenced by factors such as structure, type, format, and context. In this context, the study identifies data nature issues such as *high dimensionality, lack of structure, mixed-frequency data, non-stationarity, non-Gaussian distribution, nonlinearity, non-heterogeneity, and heteroscedasticity*.

*Data quality issue*

The incompleteness of data, often referred to as missing, fragmented, or insufficient data, may arise due to unstandardized data collection or data loss during storage or transfer. To tackle this issue, Liu et al. (2023) recommended leveraging a multitude of meteorological data prediction sources and employing data assimilation techniques. Shi et al. (2023) suggested that for a small amount of missing data (<5%), it can be safely disregarded, while for a significant amount of missing data (>60%), deletion of the dataset without utilization is advised. For intermittently short-term data loss, interpolation techniques can be applied. In cases of continuous long-term data loss, they also recommended analyzing the correlation between missing data variables and other complete variables, followed by filling in the missing data. Furthermore, Chuo et al. (2022) proposed the use of data imputation methods, involving the replacement of missing or incomplete data values with estimated or predicted values. In situations where there are insufficient samples in the collected dataset, they suggested considering the application of additional sensors or generating virtual data using generative adversarial networks (GANs). Tripathi et al. (2021) also acknowledged the appropriateness of deletion when necessary. Alternatively, imputation techniques, such as replacing with average/most frequent values, multiple imputations, expectation maximization algorithm, and classification/regression trees, can be employed. Zhang & Gao (2021) emphasized the importance of data imputation as well. Moreover, they suggest for time series data, recurrent neural networks can be applied, and for image convolutional data, neural networks or hybrid approaches are recommended. While data augmentation serves to tackle data incompleteness, it is not without its challenges. Generated images through data augmentation may exhibit high similarity, and an overreliance on this technique may introduce the multicollinearity problem (Liu et al., 2023). Strategies to address these issues include acquiring more image data, regulating the extent of data augmentation, and leveraging emerging data collection technologies such as crawlers, big data, and image data augmentation based on GANs.

Data imbalance is a very common issue in the context of ML and data analysis (Abrasaldo et al., 2024; Aldoseri et al., 2023; Strielkowski et al., 2023b; Chuo et al., 2022; Zhang & Gao, 2021), refers to a situation where the distribution of classes in a dataset is not uniform. To address this issue, two types of approaches are suggested. Firstly, at the data level, Abrasaldo et al. (2024) proposed employing data resampling techniques, such as random under-sampling. Additionally, Aldoseri et al. (2023) emphasized the importance of meticulous attention to data collection and pre-processing. Chuo et al. (2022) advocated for the use of under/over-sampling methods, while Zhang & Gao (2021) suggested data interpolation, exemplified by the Synthetic Minority Over-sampling Technique. Another approach involves the use of generative models, such as the Generative Model, a type of machine learning model designed to produce new data instances resembling a given training dataset, or GANs. GANs consist of two neural networks – a generator that creates synthetic data samples and a discriminator that evaluates the authenticity of a given sample. Secondly, at the algorithm level, Aldoseri et al. (2023) underscored the significance of understanding the principles of algorithmic fairness, fairness metrics, explainable AI, and interdisciplinary collaboration. Strielkowski et al. (2023b) posited that deep learning algorithms may need to generalize more effectively to underrepresented classes. Furthermore, Chuo et al. (2022) suggested adjusting the weight values of minor classes during the training process as a means to address data imbalance effectively.

Data inaccuracy is another issue. In dealing with inaccurate values, Chuo et al. (2022) recommended several strategies, including reducing sampling frequency, filling empty sections using interpolation methods, and ensuring synchronization in the data processing stage when dealing with different sampling rates. Addressing data noise, Zhang & Gao (2021) suggested employing data denoising techniques. Specifically, for projection-based noise, the recommendation is to utilize local geometric projection. For frequency-based noise, employing methods such as empirical model decomposition or wavelet transform is suggested. In the case of noise-assisted noise, they advised using stochastic resonance, while for data-driven/hybrid noise, the proposed approach involves utilizing generative prior and unrolled optimization.

Data outliers are observations or data points that deviate significantly from the overall pattern or distribution of the dataset. In the context of Blast Furnace ironmaking, Shi et al. (2023) suggested that outliers may arise from manual input errors, sensing device failures, or abnormal furnace states. They propose employing statistical methods, such as box plots and three-sigma analysis, for outlier identification. Additionally, ML methods like clustering and isolated forests can be utilized. From a management perspective, process-oriented approaches, such as implementing operating guidelines, are recommended. For outlier detection, Zhang & Gao (2021) advocated the use of Autoencoder at the data level. At the model level, techniques such as Probabilistic Neural Network, Temperature Scaling, and Input Perturbation are suggested. These methods enhance the ability to identify and manage outliers in the Blast Furnace ironmaking process.

Mislabeling is also a common issue. Low label accuracy and mislabeling errors can cause inefficient interpretability of datasets. In terms of label accuracy, Zhang et al. (2023b) indicated that the experimental results such as mechanical properties and computational results such as thermal distribution obtained from the finite element method model are more promising than manual labeling. Zhang & Gao (2021) introduced various techniques for data annotation. Specifically, in the realm of image annotation, suggested methods include Fully Convolutional Network, U-Net, and Mask Region-Based CNN. In the domain of natural language processing, recommended techniques involve Word Embedding and Bidirectional Encoder Representations from Transformers. Regarding manual data annotation, it can pose a challenge due to its significant time and energy consumption. Aldoseri et al. (2023) recommended techniques such as active learning, weak supervision, and transfer learning to alleviate this burden. Zhang et al. (2023b) proposed the utilization of annotation tools.

*Data nature issue*

The challenge of high dimensionality, often referred to as the "curse of dimensionality" (Sengupta, 2023), is a prevalent issue in machine learning. This arises from the complexity and high dimensionality of data, leading to the utilization of numerous predictors, potentially resulting in overfitting and diminished accuracy (Akhtar et al., 2023). Compounding the problem is the scenario where the data dimension is large, yet the sample size is very small (Roy et al., 2022). To mitigate these challenges, recommended approaches include the use of feature selection techniques and dimensionality reduction methods, such as PCA (Akhtar et al., 2023; Tripathi et al., 2021). Specifically addressing the spatial dimension of load data and the temporal and spatial variability in load data, suggestions involve employing spatial–temporal models, clustering, and spatial interpolation techniques (Akhtar et al., 2023). These strategies aim to enhance the handling of high-dimensional and complex data, contributing to more effective and accurate machine learning outcomes.

Baloch et al. (2023) highlighted the challenge of undefined structure in big data and recommends addressing it through Information Extraction (IE) technologies and advanced analytics such as NLP, ML, and DL.

Shi et al. (2023) pointed out the complexity introduced by mixed-frequency data, making future data processing and model development challenging. To tackle this issue, suggestions include converting high-frequency data to low-frequency by using methods like averaging, summing, or selecting the latest value; converting low-frequency data to high-frequency through copying; developing mixed-frequency data models using mixed data sampling and mixed frequency vector autoregression.

Certain statistical characteristics of data can pose challenges as well. One such characteristic is non-stationarity, which refers to variations in the statistical properties of time series data over time (Akhtar et al., 2023; Roy et al., 2022; Field et al., 2021). To address this issue, methods such as employing time-varying models like autoregressive moving average models or utilizing DL models such as DNNs have been suggested (Akhtar et al., 2023). Additionally, Field et al. (2021) proposed incorporating attention or memory mechanisms into the model development process to handle temporal changes and geographical variations effectively. Akhtar et al. (2023) also noted the Non-Gaussian and heavy-tailed nature of load data, recommending the use of robust Short-Term Load Forecasting (STLF) models and distributional STLF models. Addressing nonlinearity is identified as another challenge (Akhtar et al., 2023; Guo et al., 2023). Proposed methods include the application of non-parametric models like decision trees or random forests, kernel-based models such as kernel regression or support vector machines (Akhtar et al., 2023). Evaluation strategies involve assessing performance against a gold standard, analyzing performance based on experimental groups, ML classifiers, testing against expert-labeled test sets, and integrating statistical data-driven methods with ML-based approaches (Guo et al., 2023). Yüce et al. (2022) mentioned the challenge of non-heterogeneous data across various models, which indicates a lack of diversity or variability among the components. Their suggestion involves connecting previously unconnected data sources. The issue of data heteroscedasticity is highlighted in (Tripathi et al., 2021), signifying a statistical phenomenon where the variability of the dependent variable is not constant across different levels or values of an independent variable. To address this concern, they recommend employing residual analysis, statistical tests, and alternative models such as weighted least squares.

Table 9. Data issues and methods in the Data Pre-processing stage

| Data issue | Specific issue | Method | AI technology | Domain |
|---|---|---|---|---|
| **Incompleteness (data quality issue)** | 1. Data incomplete (Liu et al., 2023) | Considering as many predictions meteorological data sources as possible; using data assimilation techniques | ML, DL | Wind power - prediction, forecasting |
| | 2. Missing entries in heterogeneous data sets (Roy et al., 2022) | / | Supervised machine learning | Healthcare and biomedical research |
| | 3. Missing data, 1) Small amount of missing data （<5%） or significant amount of missing data (>60%), 2) Intermittently short-term loss of data, 3) Continuously long-term loss of data (Shi et al., 2023) | 1) Ignore or delete, 2) Interpolation techniques, 3) analyzing the correlation between the missing data variables and other complete variables and then filling in the missing data | ML | Blast furnace ironmaking |
| | 4. Fragmented data (Chuo et al., 2022) | Data imputation methods | Various | Machining operations |
| | 5. Insufficient samples of the collected data set (Chuo et al., 2022) | Applying more sensors or generating virtual data using GAN | Various | Machining operations |
| | 6. Incomplete and Missing Data (Tripathi et al., 2021) | Deletion; replacement with average/most frequent values, multiple imputations, and expectation maximization algorithm, classification/regression trees for imputation | ML | Production and Manufacturing |
| | 7. Missing or incomplete data (Zhang & Gao, 2021) | Data imputation; For time series data: Recurrent neural network; For image convolutional: neural network, hybrid approach | DL | Smart manufacturing |
| | 8. The generated images by data augmentation may have | Gather more image data; control the level of data | DL | Mining industry, sorting and |

| | | high similarity; excessive reliance on data augmentation may lead to the multicollinearity problem (Liu et al., 2023) | augmentation; use emerging data collection technologies such as crawler, big data, and GAN-based image data augmentation | | optimal utilization of minerals |
|---|---|---|---|---|---|
| **Imbalance (data quality issue)** | | 1. Imbalanced input dataset (Abrasaldo et al., 2024) | Data resampling. e.g., random under sampling | ML | Geothermal energy industry |
| | | 2. Data bias and unfairness (Aldoseri et al., 2023) | Pay attention to data collection and pre-processing, Algorithmic Fairness, Fairness Metrics, Explainable AI, Interdisciplinary Collaboration | ML | / |
| | | 3. Unbalanced datasets (Strielkowski et al., 2023b) | DL algorithms may need to generalize better to underrepresented classes | DL | Electric power systems |
| | | 4. Significant difference in the amount of data between classes (Chuo et al., 2022) | Under/over-sampling of data (approaches at data level) or adjusting the weight values of minor classes in training process (approach at algorithm level) | Various | Machining operations |
| | | 5. Data imbalance (Zhang & Gao, 2021) | Data interpolation: Synthetic minority over-sampling technique; Generative model Variational autoencoder; GANs | DL | Smart manufacturing |
| **Inaccuracy (data quality issue)** | | 1. Inaccurate values measured and intermittent sampling delay (Chuo et al., 2022) | Reducing the sampling frequency, filling the empty section using interpolation methods, and synchronization in the data processing stage for different sampling rates | Various | Machining operations |
| | | 2. Data noise (Zhang & Gao, 2021) | Data denoising. Projection-based: Local geometric projection; Frequency-based: Empirical model decomposition, Wavelet transform; Noise-assisted: Stochastic resonance; Data-driven/hybrid: Generative prior, Unrolled optimization. | DL | Smart manufacturing |
| **Outlier (data quality issue)** | | 1.Manual input error or sensing device failure; Abnormal furnace state (Shi et al., 2023) | Statistical methods such as box plot and threesigma; ML methods such as clustering and isolated forests; Process methods such as operating guidelines | ML | Blast furnace (BF) ironmaking |
| | | 2. Data outliers (Zhang & Gao, 2021) | Outlier detection. Data-level: Autoencoder; Model-level: Probabilistic neural network, Temperature scaling, Input perturbation | DL | Smart manufacturing |
| **Mislabeling (data quality** | | 1. Low label accuracy, mislabeling errors (Zhang et | Finite element method model | ML | Additive manufacturing |

| | | | | |
|---|---|---|---|---|
| issue) | al., 2023b) | | | |
| | 2. Inefficient interpretability of datasets (Zhang & Gao, 2021) | Data annotation. Image annotation: Fully convolutional network, U-Net, Mask region-based CNN; Natural language processing: Word embedding, Transformer, BERT | DL | Smart manufacturing |
| | 3. Burden of manual data annotation (Aldoseri et al., 2023) | Active learning, weak supervision, transfer learning | ML | / |
| | 4. Time consuming manual labeling process (Zhang et al., 2023b) | Annotation tool | ML | Additive manufacturing |
| **High dimensionality (data nature issue)** | 1. Curse of dimensionality (Sengupta, 2023) | / | Transformer architectures, DL | Healthcare |
| | 2. High-dimensional and complex nature of the data (Akhtar et al., 2023) | Feature selection techniques; dimensionality reduction techniques | ML, ANNs | Energy industry |
| | 3. Spatial dimension of load data, load data varies over time and space (Akhtar et al., 2023) | Spatial–temporal models; clustering and spatial interpolation techniques | ML, ANNs | Energy industry |
| | 4. The data dimension is large, but the sample size is very small (Roy et al., 2022) | / | SML | Healthcare and biomedical research |
| | 5. High dimensionality (Tripathi et al., 2021) | Dimensionality reduction (e.g., PCA) and feature selection methods | ML | Production and Manufacturing |
| **Lack of structure (data nature issue)** | 1. Big data lacks defined structure (Baloch et al., 2023) | Information extraction tech, and advanced analytics like NLP, ML, and DL | NLP, ML, and DL | Healthcare |
| **Mixed frequency data (data nature issue)** | 1. 1) High-frequency data are converted to low-frequency ones, 2) Low-frequency data are converted to high-frequency ones, 3) Mixed-frequency data model (Shi et al., 2023) | 1) Average or sum or latest value, 2) Copy, 3) Mixed data sampling and mixed frequency vector autoregression | ML | Blast furnace ironmaking |
| **Non-stationary (data nature issue)** | 1. Dynamic nature of load data, non-stationary data (Akhtar et al., 2023) | Time varying models, such as autoregressive moving average models; DL models, such as DNNs | ML, ANNs | Energy industry |
| | 2. Non-established and non-stationarity of data (Roy et al., 2022) | / | SML | Healthcare and biomedical research |
| | 3. Non-Stationary Data Distribution (Field et al., 2021) | Attention or Memory Mechanisms can be embedded into the model development process to handle changes over time and variations in different geographical locations | ML | Radiation oncology |
| **Non-Gaussian distribution** | 1. Non-Gaussian and heavy-tailed nature of the load data | Robust STLF models; distributional STLF models | ML, ANNs | Energy industry |
| (data nature issue) | (Akhtar et al., 2023) | | | |
| **Nonlinearity (data nature issue)** | 1. Nonlinear and non-monotonic relationships between the load data, weather data, and other predictors (Akhtar et al., 2023) | Non-parametric models, such as decision trees or random forests; kernel-based models, such as kernel regression or support vector machines | ML, ANNs | Energy industry |
| | 2. Presence of highly correlated variables; Large number of variables (Guo et al., 2023) | Evaluating performance against a gold standard, analyzing performance based on experimental groups, matching ML classifiers, and testing against expert-labeled test sets | ML | Materials science and engineering |
| | 3. Massiveness, nonlinearity, and high dimensionality of data (Guo et al., 2023) | Combining the statistical data-driven methods with the ML based methods | ML | Wind energy Infrastructure |
| **Non-heterogeneity (data nature issue)** | 1. Non-heterogenous data across different models (Yüce et al., 2022) | Connecting previously unconnected data sources | ML | Wind energy infrastructure |
| **Heteroscedasticity (data nature issue)** | 1. Data Heteroscedasticity (Tripathi et al., 2021) | Residual analysis, statistical tests, and alternative models like weighted least squares | ML | Production and Manufacturing |

**5) Data Processing**

In the data processing stage, 6 data issues have been identified, falling into the categories of lack of efficiency, complicated procedures, and model overfitting. Addressing the lack of efficiency, KHALAF et al. (2023) pointed out the absence of efficient data processing and analysis capabilities, along with a deficiency in reliable communication systems. This can be mitigated through the implementation of Edge Computing and IoT. Edge Computing and IoT contribute to reducing latency and facilitating real-time data processing and analysis. Furthermore, Aldoseri et al. (2023) emphasized the challenge of processing large datasets that demand enormous computing resources. To address this, specialized hardware like GPUs and TPUs can be employed to accelerate AI training and inference. Techniques such as model compression, pruning, and quantization are suggested for optimizing AI models, while transfer learning proves beneficial in reducing the amount of required training data and enhancing the efficiency of the training process. A challenge highlighted by Wen & Li (2022) pertained to dealing with dataset sizes to strike a balance between performance and acceptable running times. Ensuring diverse, representative, and reliable data through resource distribution becomes imperative in overcoming this challenge. Additionally, Yüce et al. (2022) underscored the necessity, at times, to use longer datasets. Therefore, the development of models with a large degree of freedom becomes crucial in such scenarios. In terms of complicated procedures, Wong et al. (2023) noted the existence of complicated testing procedures, specifically related to using ML for vaccine development. Concerning model overfitting, Strielkowski et al. (2023a) highlighted the risk of models becoming overly complex and lacking generalizability to new data. However, neither source provides specific methods to address these identified issues.

Table 10. Data issues and methods in the Data Processing stage

| Data issue | Specific issue | Methods | AI technology | Domain |
|---|---|---|---|---|
| **Lack of efficiency** | 1. Lack of efficient data processing and analysis capabilities, and lack of reliable communication systems (KHALAF et al., 2023) | Edge Computing and IoT | ML | Oil and gas industry – maintenance |
| | 2. Processing large datasets requiring enormous computing resources to (Aldoseri et al., 2023) | Specialized hardware such as GPUs and TPUs: Model compression, pruning, and quantization; Transfer learning | ML | / |
| | 3. Dealing with dataset sizes to balance performance vs. acceptable running times (Wen & Li, 2022) | Ensuring diverse, representative, and reliable data through resource distribution | AutoML | Spatial decision support systems |
| | 4. Using longer data sets (Yüce et al., 2022) | Develop models with large degrees of freedom | ML | Wind energy infrastructure |
| **Complicated procedures** | 1. Complicated testing procedures specifically related to using ML for vaccine development (Wong et al., 2023) | / | ML | Biomedical research and healthcare |
| **Model Overfitting** | 1. Models becoming overly complex and lacking generalizability to new data (Strielkowski et al., 2023a) | / | ML | Power systems sector |

**6) Data Security and Privacy**

In the realm of Data Security and Privacy, 6 issues are identified, focusing on information leakage, information misuse, and attacks. Addressing information leakage, Durlik et al. (2023) highlighted the risk of compromising data security leading to the leakage of information. To mitigate this, the following suggestions are proposed: prioritize data encryption, mandate regular security audits and penetration testing, and establish robust access controls along with rigorous authentication protocols. Additionally, Wilson et al. (2021) pointed out the risk of inadvertent release of identifiable information or identification through deductive disclosure. From a management perspective, they recommend bringing data scientists and data processing capabilities into the organization and building a trusted research environment.

Addressing information misuse, unauthorized access, breaches, and the improper use of sensitive information become concerns. Thus, it is imperative to implement robust data privacy and security measures, encompassing encryption, access restrictions, and compliance with data protection regulations (Cellina et al., 2023). Baloch et al. (2023) also referenced breaches and recommends the utilization of an unsynchronized sensor data analytics model as a potential solution.

Regarding attacks, the presence of inference attacks, model inversion, or membership inference attacks raises concerns. Solutions such as Federated training and Differential Privacy (DP) are suggested as protective measures (Aldoseri et al., 2023). Federated training involves training machine learning models across decentralized devices or servers, ensuring privacy by keeping raw data local and not shared. Differential Privacy (DP) is a privacy and security concept aimed at safeguarding individuals' sensitive information while still allowing valuable insights from aggregated data. It achieves this by introducing noise to the data, preserving statistical properties. The landscape also includes adversarial attacks, data poisoning, and model/data tampering. Addressing these threats requires the implementation of adversarial training techniques to bolster the robustness of machine learning models, monitoring, and anomaly detection techniques for identifying abnormal patterns or deviations from expected behavior. In addition, from a management perspective, compliance with Data Protection Regulations is imperative (Aldoseri et al., 2023).

Table 11. Data issues and methods in the Data Security and Privacy

| Data issue | Specific issue | Methods | AI technology | Domain |
|---|---|---|---|---|
| **Information leakage** | 1. Compromise on data security leading to information leakage (Durlik et al., 2023) | Prioritize data encryption; Mandate regular security audits and penetration testing; Establish robust access controls and rigorous authentication protocols | ML | Maritime industry, maritime operations and maintenance |
| | 2. Inadvertent release of identifiable information, or identification through deductive disclosure (Wilson et al., 2021) | Bring data scientists and data processing capabilities into the organization; Build a trusted research environment | / | Healthcare |
| **Information misuse** | 1. Unauthorized access, breaches, and misuse of sensitive information (Cellina et al., 2023) | Implementing robust data privacy and security measures, including encryption, access restrictions, and compliance with data protection | DL | Healthcare |

|  |  | regulations |  |  |
|---|---|---|---|---|
|  | 2. Breaches (Baloch et al., 2023) | Unsynchronized sensor data analytics model | NLP, ML, DL | Healthcare |
| **Attacks** | 1. Inference Attacks, model inversion or membership inference attacks (Aldoseri et al., 2023) | Federated training; DP | ML | / |
|  | 2. Adversarial Attacks, Data Poisoning, Model and data tampering (Aldoseri et al., 2023) | Adversarial training techniques; Monitoring and anomaly detection techniques; Compliance with Data Protection Regulations | ML | / |

**7) AI Technologies Adoption**

When contemplating the adoption of AI technologies, several challenges come to light. 5 specific issues have been identified, which can be categorized as cost issue, reusability issue, and interpretability issue. In terms of cost, KHALAF et al. (2023) argue that installing the required sensors and infrastructure for data collection can be costly, thus a Cost-Benefit Analysis should be conducted first. Durlik et al. (2023) further underscored the financial burden associated with implementing advanced data analytics, encompassing expenses related to hardware, software, infrastructure, and staff training or hiring.

In terms of reusability, Wen & Li (2022) highlighted challenges tied to updating existing models with new data while ensuring performance consistency and reproducibility, without, however, providing specific solutions.

On the front of interpretability, Wen & Li (2022) raised challenges related to understanding why models perform as they do and explaining the reasons behind specific model actions. Additionally, Xu et al. (2022) expressed doubts about the reliability of Deep Neural Networks and recommends employing Concept Drift Detection, Uncertainty Estimation, and Out-of-Distribution Detection when deploying models in real-world and dynamic environments to ensure their reliability. Concept Drift Detection is aimed at identifying when the statistical properties of the target variable change, allowing for timely model adaptation. Uncertainty estimation involves quantifying the uncertainty linked to predictions made by ML models. Out-of-distribution detection focuses on identifying instances where input data significantly deviates from the distribution on which the model was trained.

Table 12. Data issues and methods in AI Technologies adoption stage

| Data issue | Specific issue | Methods | AI technology | Domain |
|---|---|---|---|---|
| **Cost** | 1. Implementing AI technologies, installing the required sensors and | Cost-Benefit Analysis | ML | Oil and gas industry – |

| | infrastructure for data collection can be costly (KHALAF et al., 2023) | | | maintenance |
|---|---|---|---|---|
| | 2. Implementing advanced data analytics can be costly, including hardware, software, infrastructure, hiring or training staff (Durlik et al., 2023) | / | ML | Maritime industry, maritime operations and maintenance |
| **Reusability** | 1. Challenges on updating existing models with new data, while maintaining performance consistency and achieving reproducible solutions (Wen & Li, 2022) | / | AutoML | Spatial decision support systems |
| **Interpretability** | 1. Challenges on Understanding why models perform better or worse, and explaining the reasons behind certain model actions (Wen & Li, 2022) | / | AutoML | Spatial decision support systems |
| | 2. Doubts on Deep Neural Networks Reliability (Xu et al., 2022) | Concept Drift Detection, Uncertainty Estimation, Out of Distribution Detection | DL | Smart manufacturing |

## 5. Data management framework for systematically handling data issues in industrial AI

Building upon the results, this section presents a data lifecycle management framework for handling data issues in industrial AI, aiming to answer the RQ3: How can these data issues be systematically resolved?

### 5.1 Data usability evaluation in various AI algorithms

It is important to recognize that different AI algorithms have different purposes and diverse data requirements, and one dataset may not be suitable for all AI algorithms. Hence, implementing AI involves estimating data usability across various AI algorithms. As mentioned in Section 4.2, regarding AI technologies and techniques, the majority of articles center around ML. Therefore, we will focus on the data requirements of different ML algorithms to emphasize different considerations when handling data issues. According to a prominent article, ML algorithms can be primarily categorized into four types: Supervised learning, Unsupervised learning, Semi-supervised learning, and Reinforcement learning (Sarker, 2021), as depicted in Figure 5. We outline the specific data requirements based on the classification of ML algorithms.

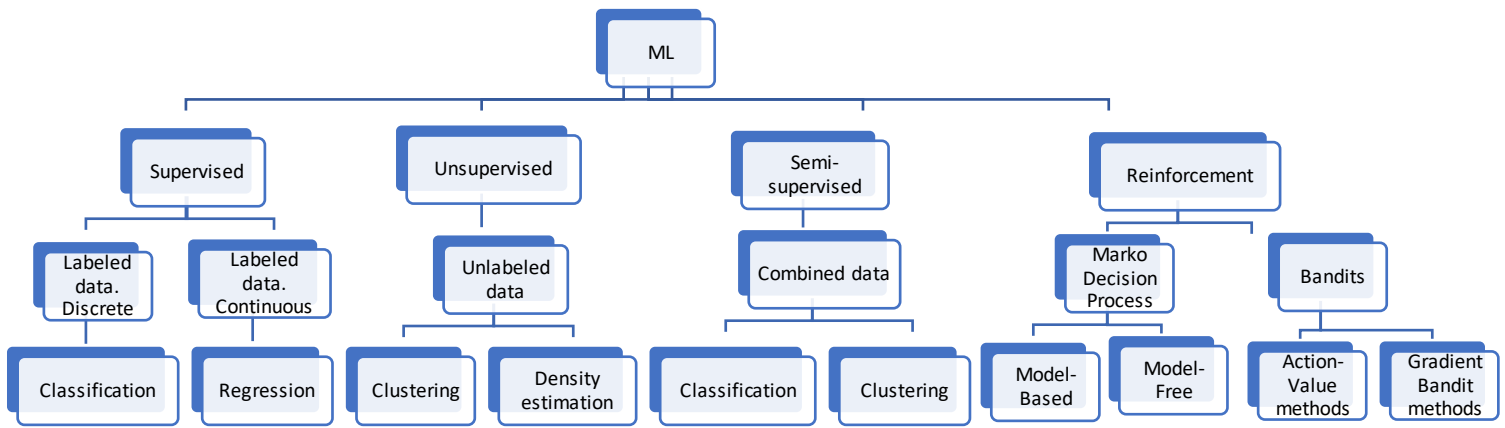

Figure 5. Categories of machine learning algorithms. Adapted from (Sarker, 2021; Zhang & Yu, 2020)

*Supervised learning*

Supervised learning utilizes labeled training data and a set of training examples to deduce a function. The primary tasks in supervised learning are classification, which segregates the data, and regression, which fits the data. It is important to have a well-structured and representative labeled dataset. Both the quality and quantity of the data significantly impact the model's performance and its ability to generalize to new, unseen data.

The key elements of the data requirement for classification in supervised learning include: 1) Input Features (X): These are the features used for making predictions or classifications; 2) Output Labels (Y): Each input sample is associated with a corresponding output label, distinguishing between classes; 3) Labeled Dataset: A dataset containing input-output pairs used for training the model; 4) Sufficient Data Variety: The dataset should encompass diverse scenarios relevant to the problem, aiding generalization; 5) Adequate Data Size: The dataset should be sufficiently large to capture underlying patterns without being excessively large, which could increase computational costs; and 6) Balanced Classes: Having a balanced distribution of examples across different classes is beneficial for training unbiased models.

In contrast to classification, where the target variable signifies discrete classes, regression involves a continuous target variable that represents a numerical value. The objective in regression is to predict a continuous output rather than categorizing samples into specific classes. Despite this distinction, the data requirements for regression are generally similar to those of classification algorithms.

*Unsupervised learning*

Unsupervised learning involves the analysis of unlabeled datasets, operating without explicit supervision or labeled output. In unsupervised learning, the system tries to learn the patterns, relationships, or structures inherent in the data without being given specific target values to predict. The most common unsupervised learning tasks are clustering, density estimation, feature learning, dimensionality reduction, finding association rules, anomaly detection, etc.

In clustering tasks, success depends on the inherent structure of the data and the appropriateness of the chosen algorithm for the given dataset. The key elements of the data requirement for clustering include: 1) Feature Set (X): The dataset should consist of a set of features that describe the characteristics of each data point; 2) Homogeneous Data Distribution: The dataset should ideally represent a somewhat homogeneous distribution of data, where clusters can be naturally identified; 3) Sufficient Data Size: Having a sufficiently large dataset is beneficial. However, the appropriate size of the dataset depends on the complexity of the data and the clustering algorithm used.

Density estimation in unsupervised learning involves estimating the probability density function of a dataset, providing insights into the underlying distribution of the data. The data requirements for density estimation include: 1) Unlabeled Data; 2) Continuous Data; 3) Representative Sample: The dataset should be a representative sample of the population or system being modeled. The quality and accuracy of the density estimation depend on how well the dataset captures the true underlying distribution; 4) Sufficient Data Size: Smaller datasets may lead to higher uncertainty in estimating the density, especially in areas with fewer observed data points; 5) Appropriate Feature Representation: The features used for density estimation should be appropriate for capturing the essential characteristics of the data distribution.

*Semi-supervised learning*

Semi-supervised learning can be defined as a hybridization of the above-mentioned supervised and unsupervised methods. It involves training a model using a combination of labeled and unlabeled data. This type of learning can be particularly useful when obtaining labeled data is costly or time-consuming. Both the labeled and unlabeled data should be representative samples of the underlying distribution. The distribution of labeled and unlabeled examples should be balanced appropriately based on the available resources and the goals of the semi-supervised learning task. The labeled and unlabeled data should be relevant to the specific task at hand. The semi-supervised learning approach is most effective when the labeled and unlabeled examples share similarities, allowing the model to leverage the unlabeled data for improved performance on the labeled task.

*Reinforcement learning*

Reinforcement learning is a type of ML where an agent learns to make decisions by interacting with an environment. The data requirements for reinforcement learning are distinct from those of supervised and unsupervised learning. It requires agent and environment interaction, environment dynamics, state representation, action space, reward signal, trajectories or episodes. In addition, the data should reflect a balance between exploration and exploitation.

*Deep learning*

As DL garners increasing attention, it is noteworthy to emphasize the distinctive data requirements associated with DL. DL is a subfield of ML, and it encompasses various types of learning paradigms, including supervised, unsupervised, and semi-supervised learning. DL models often require large and diverse datasets to learn complex patterns and generalizations. Table 13 summarizes the different data requirements for various ML algorithms.

Table 13. Different Data requirements in various ML algorithms

| ML algorithms | Task | Data requirement |
| --- | --- | --- |
| **Supervised learning** | Classification | Input features (X); Output labels (Y); Labeled dataset; Sufficient data variety; Adequate data size; Balanced classes |
| | Regression | Similar with classification, yet the target variable is continuous and represents a numerical value |
| **Unsupervised learning** | Clustering | Unlabeled dataset, Feature set (X); Homogeneous data distribution; Sufficient data size |
| | Density estimation | Unlabeled data; Continuous data; Representative sample; Sufficient data size; Appropriate feature representation |
| **Semi-supervised learning** | Classification, Clustering | Labeled data; Unlabeled data; Representative sample; Balance between labeled and unlabeled Data; Task relevance |
| **Reinforcement learning** | Reward or penalty | Agent and environment interaction; Environment dynamics; State representation; Action space; Reward signal, Trajectories or episodes; Balance between exploration and exploitation |
| **DL** | | Large and diverse datasets |

**5.2 Data management framework**

Through the synthesis of all the analyses, this section formulates a comprehensive framework for addressing data issues in industrial AI. This framework is structured into three integral parts. The initial part encompasses the data usability evaluation. Before delving into the data, we must first evaluate its usability. The initial step involves understanding the business or operational problems at hand, contemplating the purpose of utilizing AI, and specifying the tasks involved. This thoughtful process aids in better identifying potential AI algorithms and gaining insight into the required data. Subsequently, we embark on exploring the existing data to comprehend its type, features, distribution, patterns, characteristics, size, complexity, and so on. Preliminary data exploration enables a basic assessment of whether the available data can address the problem at hand and be utilized for potential AI models. Finally, we evaluate the usability of the data for specific AI models, considering the diverse requirements outlined in Section 5.1. Different AI algorithms have distinct data requirements, e.g., small datasets may not be suitable for DL. Once we ascertain that data usability is satisfactory, we can then contemplate potential data issues, e.g., when using supervised learning models, we might have to deal with mislabeling issues, and when using clustering algorithms, we might encounter non-heterogeneity issue.

The 72 data issues identified are categorized into 29 types and further classified based on the stage of the data lifecycle, as outlined in the second part of the framework. This categorization serves as a map for an overview of all the data issues, also providing a clear understanding of when and where a specific data issue might occur. Simultaneously, methods for addressing each data issue are presented in Section 4.3, offering a valuable reference for researchers and practitioners.

The final part of the framework presents corresponding methods categorized by data lifecycle, and from both a managerial and a technical standpoint. From a managerial viewpoint, for instance, during the data source and collection phase, it is recommended to foster collaboration on data collection and establish data requirements and standardization. In the data integration and interoperation stage, it becomes essential to catalyze joint development initiatives, bringing stakeholders together. Meanwhile, the technical perspective involves a range of elements like techniques, methods, tools, hardware, systems, and more, all aimed at addressing different data-related challenges. It is important to note that fitting all the methods into one chart is impossible; thus, we present only some examples of the methods addressing data issues in each stage of the data lifecycle. The detailed methods are shown in Tables 6 to 12. The framework is illustrated in Figure 6.

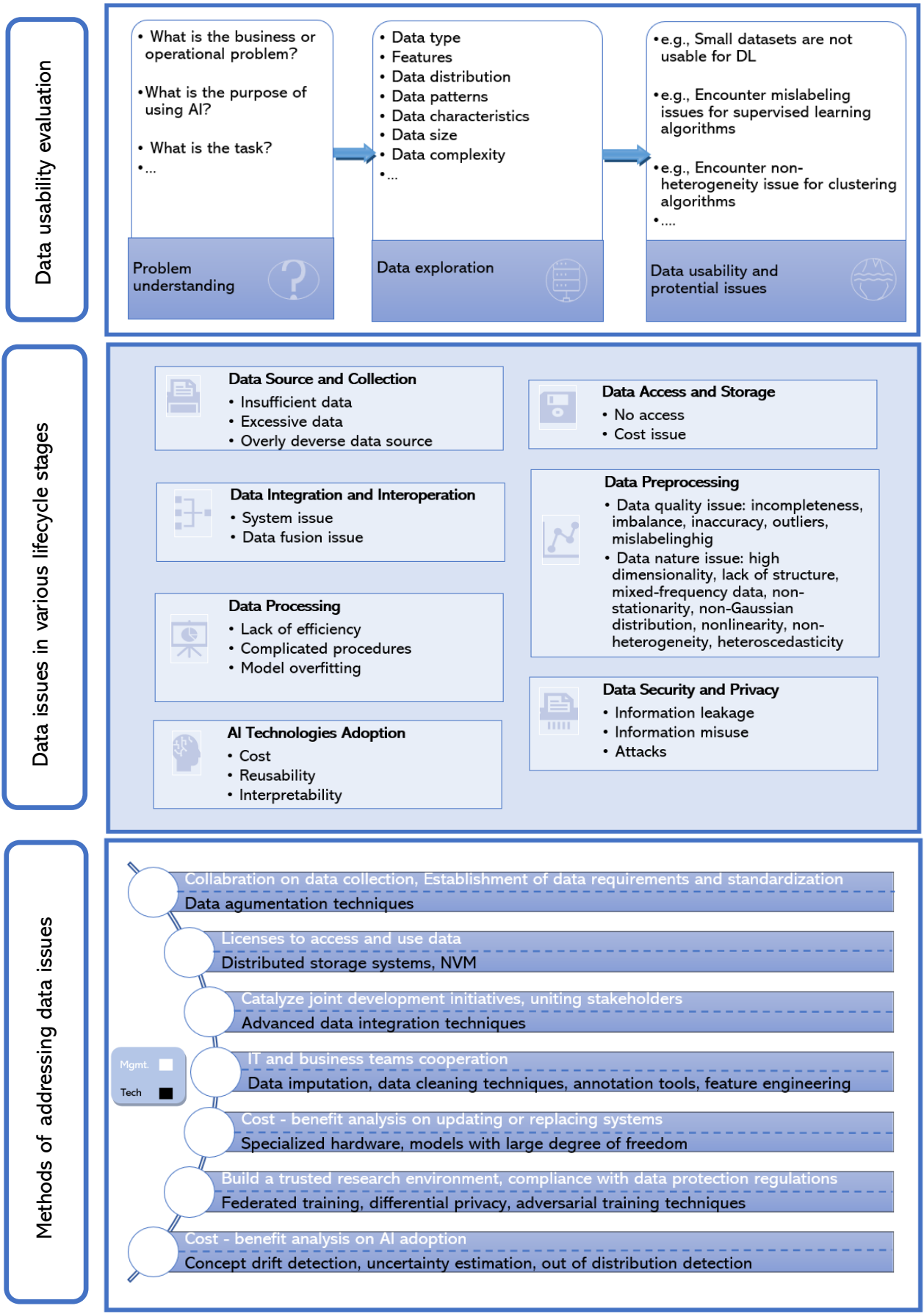

Figure 6. Data management framework for systematically handling data issues in industrial AI

# 6. Future research on data management in industrial AI

As highlighted earlier, data is essential for making actual progress in AI advancement. It is imperative for future research to prioritize the discussion of data-related topics. A data forum can be established to provide professionals with a platform to convene, exchange insights, best practices, and innovative ideas within the realms of data science, data analytics, or data management. To guide future research topics, three pivotal points can be considered.

The first point is assessing the effectiveness of existing methods. The existing methods discussed in this paper demonstrate the capability to address certain data issues. However, it is crucial to assess the effectiveness of these existing solutions and identify areas that still require improvement. This can be achieved by actively employing the methods in practice or conducting case studies on projects that have utilized these methods, subsequently evaluating, and comparing the results.

The second point is translating knowledge on solving data issues into industry experience. While the literature has highlighted numerous data issues and methods, there is a need to propel the industry towards confronting these data issues pragmatically. This requires developing and continuously practicing problem-solving skills to translate knowledge into practical experience. Given the privacy, confidentiality, and complexity of industry data, obtaining real data from specific companies is difficult. Therefore, leveraging open data sources, such as participating in initiatives like the PHM Data Challenge competition (Jia et al., 2018; Su & Lee, 2023), presents a valuable avenue for honing problem-solving skills.

The third point is achieving a more profound integration of AI and industry. Looking ahead, beyond addressing data issues, the future should also focus on expanding the application of AI. This necessitates a dual perspective: from the AI standpoint, exploring potential applications in various fields, and from the industry perspective, contemplating how to prepare industry comprehensively to pave the way for a more profound and widespread AI integration. For instance, the introduction of ChatGPT has highlighted the potential of large language models (LLMs) in showcasing artificial general intelligence. However, in industry, where there is a need for domain-specific knowledge, LLMs may not be ideal due to their training on general knowledge. Therefore, exploring the development of Industrial Large Knowledge Models (ILKMs) tailored for industrial systems could be a promising avenue for future research (Lee & Su, 2023).

# 7. Conclusions and limitations

This study conducts a meta-review to delve into data issues and methods within the implementation of industrial AI, 72 data issues are identified and categorized by the stages of the data lifecycle: data source and collection, data access and storage, data integration and interoperation, data pre-processing, data processing, data security and privacy, and AI technologies adoption. By doing so, this study answers two primary questions: "What specific data issues arise during the implementation of AI in industrial systems?" and "What methods are associated with addressing these issues?" Subsequently, this study analyzes the data requirements of various ML algorithms to guide the data usability evaluation for AI models. Finally, synthesizing all the analyses, this study proposes a data management framework for addressing data issues in industrial AI, answering the question "How can data issues be systematically resolved?". This study suggests three future research directions: 1) assessing the effectiveness of existing methods, 2) translating knowledge on solving data issues into industry experience through practical exercises, and 3) achieving a more profound integration of AI and industry. This study offers valuable insights in three areas: identifying research gaps in industrial AI, outlining practical data guidelines, and providing a comprehensive data management framework. These resources assist professionals in enhancing data usability and usefulness in industrial AI implementation.

The study acknowledges its limitations, as the literature gathered through the meta-review method may not encompass all data issues and methods. The analysis and categorization of data issues heavily rely on the availability and quality of literature sources, potentially introducing biases or gaps that may have influenced the identification and categorization process. Moreover, despite efforts to systematically analyze data issues and methods, the study is inherently subject to limitations associated with the meta-review methodology, including selection bias, publication bias, and the potential for researcher interpretation. Scholars interested in conducting a more exhaustive exploration can consider employing alternative methods such as NLP technique to automate the extraction and analysis of data-related information from a vast array of textual sources.

**Author contributions**

Xuejiao Li: Methodology, investigation, data curation, writing – original draft; Cheng Yang: writing – review & editing, project administration; Charles Møller: writing – review & editing, funding acquisition; Jay Lee: conceptualization, methodology, supervision.

**Acknowledgement**

This study is supported by MADE FAST (MADE -- Flexible, Agile, and Sustainable production enabled by Talented employees) Project (470100), Jiangxi Double Thousand Plan, and "Strengthening the digitalization of businesses in Eastern Europe – a micro and macro-level approach" funded by the European Union – NextGenerationEU project and the Romanian Government.

**Conflict of Interests**

The authors declare that there is no conflict of interests regarding the publication of this article.

**Declaration of Generative AI and AI-assisted technologies in the writing process**

Statement: During the preparation of this work the first author used ChatGPT in order to improve readability and language. After using this tool, the authors reviewed and edited the content as needed and take full responsibility for the content of the publication.